\documentclass[twoside,leqno,twocolumn]{article}
\usepackage[letterpaper]{geometry}
\usepackage{graphicx}
\usepackage{ltexpprt}
\usepackage{hyperref}
\usepackage{amsmath,amsfonts}
\usepackage{amssymb}
\usepackage{algorithmic}
\usepackage{algorithm}
\usepackage{array}
\usepackage{xspace}
\usepackage{textcomp}
\usepackage{stfloats}
\usepackage{url}
\usepackage{verbatim}
\usepackage{graphicx}
\usepackage{cite}
\usepackage{dsfont}
\usepackage{dsfont}
\usepackage{subcaption}
\usepackage{enumitem}
\usepackage{fontawesome}
\usepackage{ragged2e} 
\usepackage{textcomp}
\usepackage{booktabs,makecell, multirow, tabularx}

\makeatletter
\renewcommand*{\@fnsymbol}[1]{\ensuremath{%
    \ifcase#1\or \dagger\or \ddagger\or
    \mathsection\or \triangle \or \diamond \or \star \or \dagger\dagger
    \or \ddagger\ddagger \else\@ctrerr\fi}}
\makeatother

\begin{document}

\title{Tensorized Hypergraph Neural Networks}
\author{Maolin Wang \footnote{City University of Hong Kong \\ \quad\quad \{morin.wang@my., yzhen8-c@my., xianzhao@\}cityu.edu.hk  
}  \footnote{Antgroup \\ \{nanxiao.zy, chenyi.zcy\}@antgroup.com
} 
\and
Yaoming Zhen$^{\dagger}$ \and
Yu Pan\footnote{Harbin Institute of Technology Shenzhen \\ \{iperryuu, Zenglin\}@gmail.com} \and
Yao Zhao$^{\ddagger}$ \and
Chenyi Zhuang$^{\ddagger}$ \and
Zenglin Xu$^{\mathsection}$ \footnote{Pengcheng Laboratory} \and
Ruocheng Guo\footnote{ByteDance Research rguo.asu@gmail.com} \and
Xiangyu Zhao$^{\dagger}\footnote{Xiangyu Zhao is the corresponding author}$}

\date{}
\providecommand{\keywords}[1]
{
  \textbf{Keywords:} #1
}
\maketitle

\fancyfoot[R]{\scriptsize{Copyright \textcopyright\ 2024 by SIAM\\
Unauthorized reproduction of this article is prohibited}}

\begin{abstract}
Hypergraph neural networks (HGNN) have recently become attractive and received significant attention due to their excellent performance in various domains.
However, most existing HGNNs rely on first-order approximations of hypergraph connectivity patterns, which ignores important high-order information.
To address this issue, we propose a novel adjacency-tensor-based \textbf{T}ensorized \textbf{H}ypergraph \textbf{N}eural \textbf{N}etwork (THNN). THNN is a faithful hypergraph modeling framework through high-order outer product feature message passing and is a natural tensor extension of the adjacency-matrix-based graph neural networks. The proposed THNN is equivalent to a high-order polynomial regression scheme, which enables THNN with the ability to efficiently extract high-order information from uniform hypergraphs.  Moreover, in consideration of the exponential complexity of directly processing high-order outer product features, we propose using a partially symmetric CP decomposition approach to reduce model complexity to a linear degree. 
Additionally, we propose two simple yet effective extensions of our method for non-uniform hypergraphs commonly found in real-world applications.
Results from experiments on two widely used {hypergraph datasets for 3-D visual object classification} show the model's promising performance. 
\end{abstract}


\section{Introduction}

The rapid development of graph neural networks (GNNs, \cite{velickovic2017graph,welling2016semi,leskovec2019powerful}) greatly benefits various crucial research areas due to their extraordinary performance. Generally, a conventional GNN only allows objects to have pairwise interaction.
However, in many real-world applications, the interactions among objects can go beyond pairwise interactions and involve higher-order relationships.
For example, in brain connectivity networks~\cite{martinet2020robust},
multiple brain regions often work together in a neurological manner to accomplish certain functional tasks.
To faithfully characterize such connections, pairwise modeling in graph structure is inadequate, and it is necessary to incorporate high-order interacting information across brain regions. To articulate the correlation among multiple regions, hypergraph structures~\cite{bretto2013hypergraph,gao2020hypergraph,martinet2020robust} can be created with each vertex as a brain region and hyperedges representing the interactions among regions.

As discussed in~\cite{articlePu}, there is a clear difference between the pairwise relationship and the high order 
relationship of multiple objects. The capacity of graph structures is limited as they can only describe pairwise relationships.
Compared to graphs, a hypergraph provides significant advantages in modeling the high-order 
relationships among multiple objects in real-world data~\cite{articlePu}. For example, in the case of multi-agent trajectory prediction~\cite{xu2022groupnet}, adopting the multiscale hypergraph can extract the interactions among groups of varying sizes and performs much better than prior graph-based methods which can solely describe pairwise interactions. 
 Hypergraphs have also recently been used in a variety of other
data mining
tasks such as classification~\cite{feng2019hypergraph}, community detection~\cite{zhen2022community}, and item matching~\cite{liao2021hypergraph}.

In these applications, the majority of hypergraph neural network architectures 
are based on the Chebyshev formula for hypergraph Laplacians proposed by HGNN~\cite{feng2019hypergraph}. These neural hypergraph operators can be seen as constructing a weighted graph and thus can utilize the off-the-shelf graph learning models (e.g., GCN). \textbf{However, most of these methods are incapable of learning higher-order information since they only make use of the first-order approximation}, e.g., the clique expansion~\cite{gao2020hypergraph} of a hypergraph. 
In order to better characterize high-order information in hypergraphs, it is natural to model the high-order interaction information of a hypergraph through high-order representation (like outer product).

Some studies~\cite{ke2019community,zhen2022community} have revealed the great success of tensor representation (like adjacency tensor~\cite{zhen2022community}) in hypergraph modeling.
\textbf{However, a general tensor based hypergraph neural network, which conducts a high-order information message passing procedure, has not yet been developed until now.}

Motivated by the two aforementioned observations, in this work, we propose a tensor based \textbf{T}ensorized \textbf{H}ypergraph \textbf{N}eural \textbf{N}etwork (THNN) to extend the adjacency matrix based graph neural networks into an adjacency tensor based framework.
THNN has high expressiveness due to its intrinsic similarity to 
a high-order outer product feature aggregation scheme~\cite{zadeh2017tensor,hou2019deep,liu2018efficient},
which can capture intra-feature and inter-feature dynamics in multilinear interaction information modeling. 
In other words, the intrinsic multilinear mathematical architecture of THNN is effective and natural in modeling high-order information, resulting in a more accurate extraction of high-order interactions.

Furthermore, because adjacency tensors can only be used to represent uniform hypergraphs, the straightforward THNN is incapable of handling the widely existing non-uniform hypergraphs.
Therefore, we propose two novel solutions: (1) adding a global node 
and (2) multi-uniform processing. To evaluate the performance of the proposed THNN framework, experiments on two 3-D visual object recognition datasets are performed. The experimental results show that the proposed THNN model achieves state-of-the-art performance.
In summary, our major contributions are as follows:

\begin{itemize}
[noitemsep,topsep=10pt,leftmargin=10pt]
    \item We propose, to the best of our knowledge, the first hypergraph neural network based on adjacency tensor that {comes with a message passing mechanism capturing high-order interactions in hypergraphs}. Previous hypergraph neural networks are mostly based on the first-order approximation. 
    \item {Given the fact that the naive outer product based model of high-order information suffers from exponential time/space complexity}, we propose to utilize partially symmetric
    CP decomposition to reduce time/space complexity from exponential to linear. 
    \item {To handle non-uniform hypergraphs, }we propose two simple yet effective solutions, i.e., adding a global node and multi-uniform processing, to overcome the limitation that straightforward THNN of adjacency tensor methods can only be used to model and process uniform hypergraphs.
\end{itemize}
\vspace{-2pt}

\section{Preliminaries and Background}
\subsection{Graph and Hypergraph}

A graph can be denoted by $G = (V, E)$, where $V$ is the set of
vertices and $E$ is a set of paired vertices, or edges. A graph can be represented by its adjacency matrix $\mathbf{A}\in \{0,1\}^{|V|\times |V|}$, where $|\cdot|$ denotes the set cardinality. The matrix $\mathbf{A}$ entries indicate whether two vertices in the graph are adjacent. More specifically, $\mathbf{A}_{i,j} = 1$ if $\left\{v_i, v_{j}\right\} \in {E}$ and $0$ otherwise. A hypergraph $\mathcal{G}=(\mathcal{V},\mathcal{E})$ is a generalization of a graph in which any number of vertices can be joined in one edge. $\mathcal{V}$ is the set of
vertices, and $\mathcal{E}$ is a set of vertex sets, a.k.a hyperedges. A hypergraph~(undirected) is always described by an incidence $\mathbf{H}\in\{0,1\}^{|\mathcal{V}|\times |\mathcal{E}|}$, where $|\mathcal{V}|$ is the number of vertices and $|\mathcal{E}|$ is the number of hyperedges. Specifically, $\mathbf{H}_{i,j}=1$, if $v_i \in e_j$ and $\mathbf{H}_{i,j}=0$ if $v_i \notin e_j$ . 
\begin{figure}[!ht]
\centering
\setlength{\belowcaptionskip}{-0.3cm}  \includegraphics[width=0.35\textwidth]{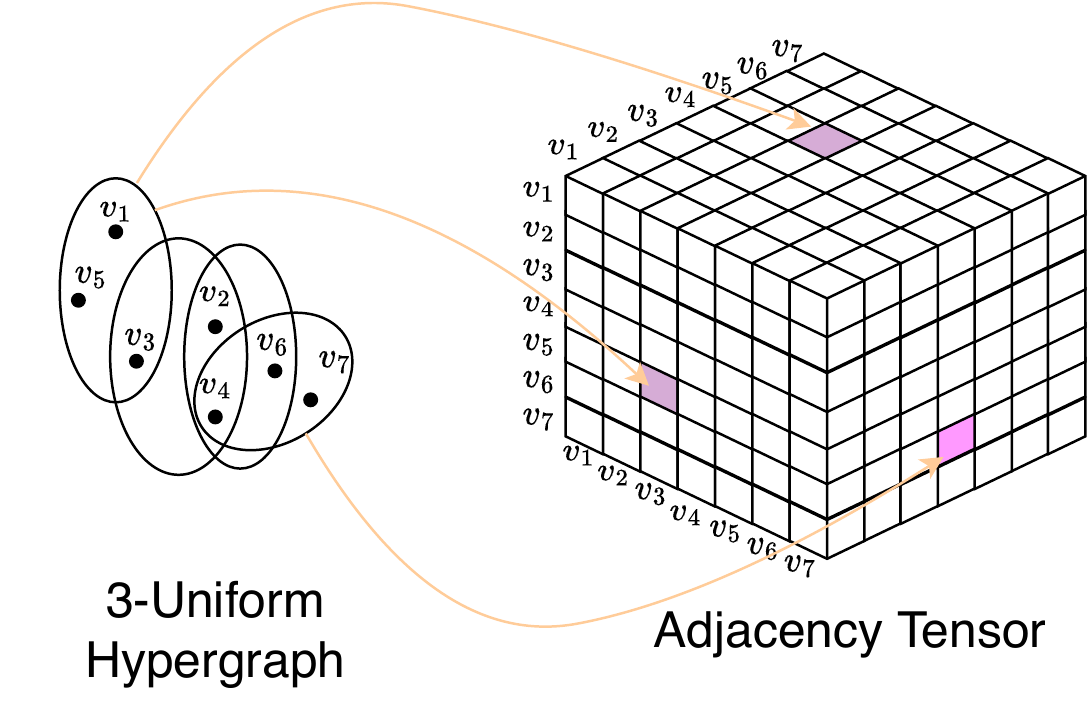}
  \caption{{An example of adjacency tensor of a 3-uniform hypergraph. In this example, the adjacency tensor of hypergraph $\mathcal{G}$ is defined as the $3$-order tensor $\mathcal{A} \in\{0,1\}^{7 \times 7 \times 7}$ with the entry $\mathcal{A}_{v_{i},v_{j},v_{k}} = 1$ if $\left\{v_{i}, v_{j}, v_{k}\right\} \in \mathcal{E}$ and 0 otherwise.}
 }
  \label{fig:adjacency}
 \end{figure}

Hypergraphs can be approximated by graphs via its clique expansion~\cite{gao2020hypergraph}. 
The clique expansion approximates the original hypergraph $\mathcal{G} = (\mathcal{V}, \mathcal{E})$ via a graph $G_{clique} = (\mathcal{V}, E_{clique})$, which reduces each hyperedge $e \in \mathcal{E}$ into a clique in $G_{clique}$. However, the clique expansion will lead to information loss~\cite{yang2020hypergraph}. The original hypergraph can not be recovered according to the adjacency matrix of clique expansion, as the hyper-dependency and high order relationship collapses into linearity~\cite{yang2020hypergraph}.

Another important way to represent hypergraphs is by \textbf{Adjacency Tensor}~\cite{zhen2022community}.
As shown in Figure~\ref{fig:adjacency}, adjacency tensor can represent uniform hypergraph where all hyperedges share the same size. The $m$-uniform hypergraph means that the sizes of all hyperedges are $m$.
For an $m$-uniform hypergraph, the adjacency tensor is defined as the $m$-order tensor $\mathcal{A} \in\{0,1\}^{n \times \ldots \times n}$ with the entry $\mathcal{A}_{i_1,...,i_m} = 1$ if $\left\{v_{i_{1}}, \ldots, v_{i_{m}}\right\} \in \mathcal{E}$ and 0 otherwise.

\subsection{Tensor Contraction}

{Tensor contraction~\cite{matthews2018high,DBLP:conf/ijcnn/WangZPXX19} means that two tensors are contracted into one tensor along their associated pairs of indices. Given two tensors ${ \mathcal{A}} \in \mathbb{R}^{I_1\times I_2\times \cdots \times I_N}$ and ${\mathcal{B}}\in\mathbb{R}^{J_1\times J_2\times  \cdots \times J_M}$, with some common modes, $I_{n_1}=J_{m_1}$, $\cdots$ $I_{n_S}=J_{m_S}$, the tensor contraction ${\mathcal{A}} \times^{(n_1,n_2\cdots,n_S)}_{(m_1,m_2\cdots,m_S) } {\mathcal{B}}$ yields a $(N+M-2S)$-order tensor ${\mathcal{C}}$.
Tensor contraction can be formulated as:
\begin{align}
     &\mathcal{C} = \mathcal{A} \times_{(j_{m_1}, j_{m_2}, \dots j_{m_S})}^{(i_{n1}, i_{n2}, \dots i_{n_S})} \mathcal{B}  \notag
\\&
    = \sum_{i_{1},i_{2},\cdots i_{N}} \mathcal{A}_{i_1,i_2,\cdots i_{n_S},*}\quad\mathcal{B}_{*,i_1,i_2,\cdots i_{n_S}}.\notag
\end{align}

The well known Mode-$N$ Product is a special case of Tensor Contraction. Given a tensor ${ \mathcal{A}} \in \mathbb{R}^{I_1\times I_2\times \cdots \times I_N}$ and a matrix ${\mathbf{B}}\in\mathbb{R}^{J_1\times J_2}$. If $J_2 = I_{n}$, then
\begin{equation}
    {\mathcal{C}}
    =  {\mathcal{A}} \times^{(n)}_{(2) } {\mathbf{B}} = {\mathcal{A}} \times_{n} {\mathbf{B}}.\notag
\label{eq:tensor_contraction}
\end{equation}

\begin{figure}[t]
\centering
 \includegraphics[width=0.47\textwidth]{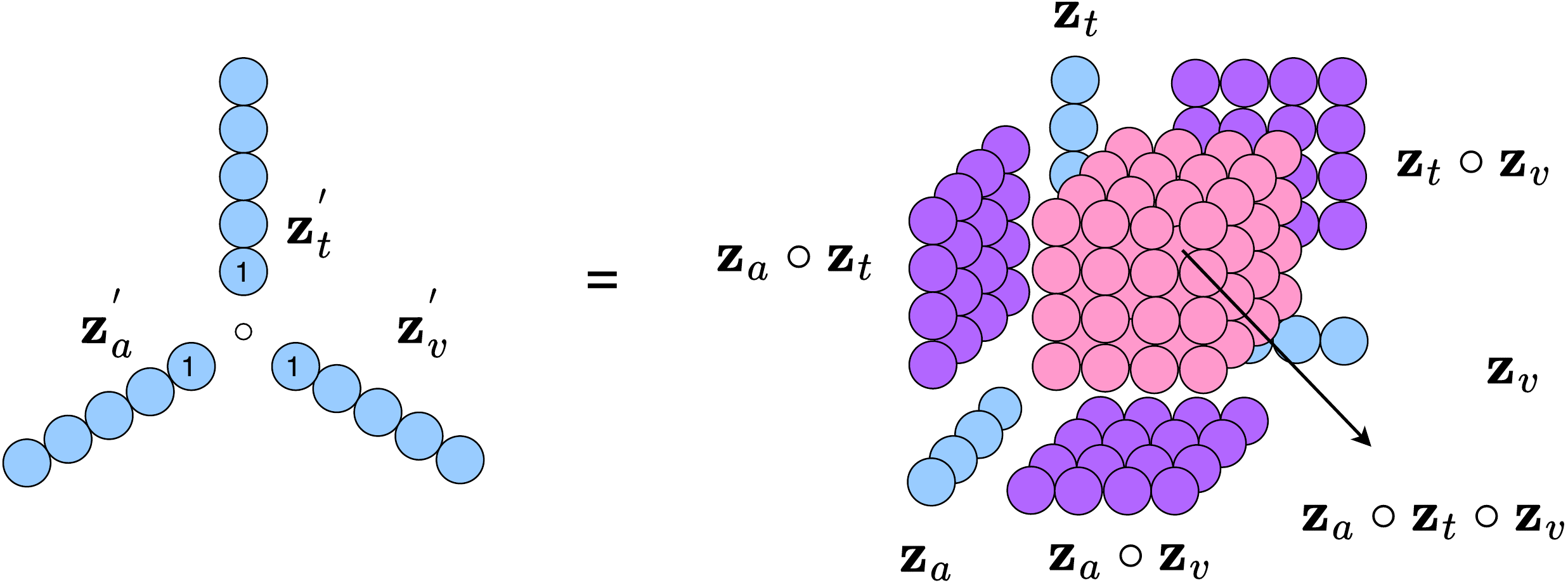}
  \caption{Illustration of concatenating 1 and
  high-order fusion~\cite{zadeh2017tensor}. Every circle corresponds to an element in a vector or tensor and $\circ$ indicates the outer product. We can concatenate a 1 to each vector, and then the outer product of vectors will introduce lower order dynamics. }
  \label{fig:fusion}
 \end{figure}

\section{Methods}

{In this section, we first analyze the widely adopted hypergraph neural network -- HGNN~\cite{feng2019hypergraph}, which uses first-order information for hypergraph representation learning. Next, we propose and analyze tensorized hypergraph neural network based on adjacency tensors. Since the straightforward THNN cannot handle more common non-uniform hypergraphs, we introduce two simple yet effective solutions: global node adding and multi-uniform processing. }

Feng et al. \cite{feng2019hypergraph} develop the classical Hypergraph Neural Networks which use truncated Chebyshev formula as hypergraph Laplacians. Given the incidence matrix $\mathbf{H} \in \mathbb{R}^{|\mathcal{V}| \times |\mathcal{E}|}$ of the hypergraph $\mathcal{G}$, its operator can be written as
\begin{equation}
\label{eq:hgcn}
\mathbf{X}^{(l+1)}=\sigma\left(\mathbf{D}_{(v)}^{-1 / 2} \mathbf{H} \mathbf{W} \mathbf{D}_{(e)}^{-1} \mathbf{H}^{\top} \mathbf{D}_{(v)}^{-1 / 2} \mathbf{X}^{(l)} \mathbf{\Theta}^{(l)}\right),
\end{equation}
where $\mathbf{W} \in \mathbb{R}^{|\mathcal{E}| \times |\mathcal{E}|}$ is a diagonal matrix to be learned and $\mathbf{\Theta}^{(l)}$ is a learnable matrix in layer $l$. $\mathbf{D}_{(v)ii} = \sum_{j=1}^{|\mathcal{E}|} W_{j j} H_{i j}$ and $\mathbf{D}_{(e)jj} = \sum_{i=1}^{|\mathcal{V}|}  H_{i j}$
are diagonal degree matrices of vertices and edges, respectively. These methods can be viewed as applying clique expansion graph that all the edges in a same clique sharing the same learnable weight to approximate the hypergraph.
Each hyperedge of size $s$ is approximated by a weighted $s$-clique.  
By analyzing the computation of Eq.~(\ref{eq:hgcn}), we denote the embedding of node $v_{i}$ in the $l+1$-th layer by $x_{v_{i}}^{(l+1)}$. The computation can be demonstrated as the following aggregation function form,
\begin{equation}
x_{v_{k}}^{(l+1)}=\sigma\bigg(\frac{1}{\sqrt{d_{v_k}}}\sum_{e_j, v_k \in e_j}  \frac{1}{d_{e_j}}\mathbf{W}_{jj} \sum_{v_i, v_i \in e_j}\frac{1}{\sqrt{d_{v_i}}} x_{v_{i}}^{l} \mathbf{\Theta}^{(l)}\bigg).
    \label{eq:hgcn2}
\end{equation}

This approach tackles information aggregation by a weighted summation of the linearly processed (via $ \mathbf{\Theta}^{(l)}$) node embeddings of neighbors in the weighted clique expansion graph. However, it is insufficient for higher-order information extraction as only the first-order linear information is considered in Eq.~\eqref{eq:hgcn2}.

\subsection{Tensorized Hypergraph Neural Network}

Eq.~(\ref{eq:hgcn2}) has revealed that classical Hypergraph Neural Networks approximate high-order information via the first-order summation. {However, higher-order information better characterizes co-occurrence relationship in hypergraph.}

In order to characterize the influence of other nodes in the same hyperedge on its high-order interaction information, for a node in a hypergraph, the most intuitive method is to use the outer product pooling~\cite{zadeh2017tensor} of the feature vectors of its neighbors.
For example, for a node $v_i$ of a hyperedge $\left\{v_i, v_{j}, v_{k}\right\}\in\mathcal{E}$ in a third-order hypergraph $\mathcal{G}$, the message of the hyperedge $\left\{v_i, v_{j}, v_{k}\right\}$
to node $v_i$ is $x_{v_j} \circ x_{v_k} \in \mathbb{R}^{I_{in}\times I_{in}}$. Similar to other graph neural networks, a trainable weight tensor can process and align features, followed by aggregating all hyperedges. 
Then, the information aggregation of node $v_i$ embedding can be represented as:
\begin{equation}
    \label{Eq:outer_product}
    x^{(l+1)}_{v_i} = \sum_{(j,k) \in N_i} x^l_{v_j} \circ x^l_{v_k} \times^{(2,3)}_{(1,2) } \mathcal{W},
\end{equation}
where $\mathcal{W} \in \mathbb{R}^{I_{in} \times I_{in} \times I_{out}}$  is the weight tensor. , where $N_i$ is the set of neighbor pairs of the node $v_i$ and $\circ$ indicates the outer product. 
The general $N$-th order form. Then, the outer product information aggregation of node $v_i$ embedding can be represented as follows:
\begin{equation}
    \label{Eq:outer_product_2}
    x_{v_{j_1}} = \sum_{(j_1,j_2,\cdots,j_{N-1}) \in N_i} (x_{v_{j_1}} \circ \cdots \circ x_{v_{j_{N-1}}}) \times^{(2,3,\cdots,N)}_{(1,2,\cdots,N-1) } \mathcal{W}  
\end{equation}
where $\mathcal{W} \in \mathbb{R}^{I_{in} \times I_{in}\cdots \times I_{out}} \text{ is the weight tensor}$.

Eq.\eqref{Eq:outer_product} and \eqref{Eq:outer_product_2} formulate the basic framework of the hypergraph neural network that utilizes high-order polynomial information directly.
Such formulation also reminds us of the polynomial regression scheme~\cite{ hsu2010microstructural}, which are highly recognized successful techniques for high-order interaction information extracting (such as multi-modality analysis~\cite{zadeh2017tensor,hou2019deep,liu2018efficient}).

\begin{figure*}[t]
\centering
 \includegraphics[width=0.8\textwidth]{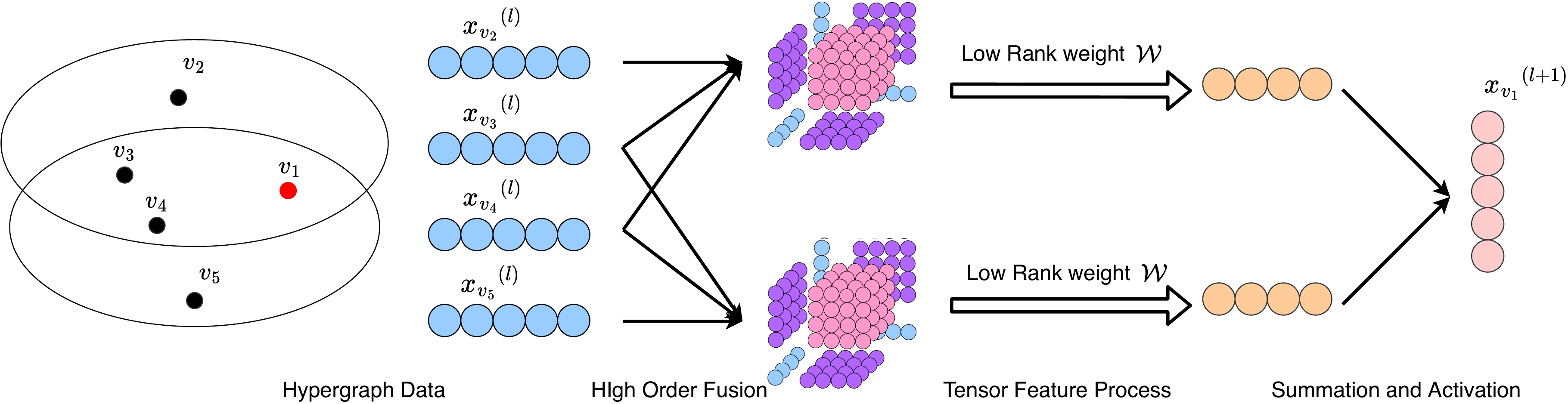}
  \caption{Illustration of THNN. THNN tries to pass the high-order interactions of neighbors in different hyperedges. The information of interactions is computed and processed via tensor operations. 
  }
  \label{fig:model}
 \end{figure*}

Similar to the representation equivalence between the aggregation scheme and adjacency matrix formulation for Graph Convolution Neural Networks~(GCN), the adjacency tensor can also be used to describe Eq.\eqref{Eq:outer_product_2}. The outer product feature aggregation can be reformulated as,

\begin{equation}
\mathbf{X}^{(l+1)}=\sigma\left(({\mathcal{A}}\times_2\mathbf{X}^{(l)} \cdots \times_N \mathbf{X}^{(l)}) \times^{(2,3,\cdots,N)}_{(1,2,\cdots,N-1) } \mathcal{W}  \right),
\label{Eq:THNN0}
\end{equation}

Similar to GCN, simple graph convolution layer without feature normalization can result in numerical instabilities because directly applying convolution layer changes the scale of feature vectors. As a consequence of this, there is a need for an appropriate level of degree-normalization~\cite{welling2016semi}. Similar to the degree-normalized adjacency matrix in GCN, we adopt the {well-known} normalizing adjacency tensor extension\cite{ouvrard2018adjacency},
\begin{equation}
    {\tilde{\mathcal{A}}_{i_{1} \ldots i_{k}}}= \begin{cases}\frac{1}{(k-1) !} \prod_{1 \leqslant j \leqslant k} \frac{1}{\sqrt[k]{d_{i_{j}}}} & \text { if }\left\{v_{i_{1}}, \ldots, v_{i_{k}}\right\} \in E \\ 0 & \text { otherwise }\end{cases}.
\label{Eq:THNN2}
\notag
\end{equation}

Thus, Eq.\eqref{Eq:THNN0} can be then reformulated as,
\begin{equation}
\mathbf{X}^{(l+1)}=\sigma\left((\tilde{\mathcal{A}}\times_2\mathbf{X}^{(l)} \cdots \times_N \mathbf{X}^{(l)}) \times^{(2,3,\cdots,N)}_{(1,2,\cdots,N-1) } \mathcal{W}  \right),
\label{Eq:THNN3}
\notag
\end{equation}
Since the size of the parameter tensor will grow exponentially with the order number, such extremely large storage and computational complexity is unacceptable. This phenomenon is known as the curse of dimensions~\cite{cichocki2014era}, and proper tensor decomposition format can effectively solve this problem. So, we decompose the weight tensor
 $\mathcal{W} \in \mathbb{R}^{I_{in} \times I_{in} \cdots \times I_{out}}$ into the following partially symmetric CP decomposition~\cite{kolda2009tensor,ni2022semidefinite} structure with the rank $R$:
$$
\mathcal{W} = \mathcal{I} \times_1 \mathbf{\Theta}^{(l)} \times_2  \mathbf{\Theta}^{(l)}\cdots \times_{N-1}  \mathbf{\Theta}^{(l)} \times_N \mathbf{Q}^{(l)T}.
$$

Using partially symmetric constraints is motivated by the assumption that the same combination of nodes in \textbf{undirected} hypergraph should result in equal output features after outer product fusion. For example, as for node pair $v_j$ and $v_k$, the weight should hold $\mathcal{W}\times_1 x_{v_j} \times_2 x_{v_k} = \mathcal{W}\times_1 x_{v_k} \times_2 x_{v_j}$. Therefore, partially symmetry constraints can address this assumption and reduce the number of parameters.
The final low-rank aggregation scheme can be represented as follows

\begin{align}
& \mathbf{X}^{(l+1)} =  
\sigma((\left(\tilde{\mathcal{A}}\times_2(\mathbf{X}^{(l)}\mathbf{\Theta}^{(l)}) \cdots \times_N(\mathbf{X}^{(l)}\mathbf{\Theta}^{(l)})\right) 
\notag
\\&
\times^{(2,3,\cdots,N)}_{(1,2,\cdots,N-1) }  \mathcal{I})  \mathbf{Q}^{(l)T})\notag
\end{align}

where $\tilde{\mathcal{A}} \in \mathbb{R}^{|\mathcal{V}| \times |\mathcal{V}| \cdots \times |\mathcal{V}|}$, $\mathbf{X} \in \mathbb{R}^{|\mathcal{V}|\times I_{in}}$, $ \mathcal{I} \in \mathbb{R}^{R \times R \cdots \times R} $ is the identity tensor, $\mathbf{\Theta}^{(l)} \in \mathbb{R}^{I_{in}\times R}$, and $\mathbf{Q}^{(l)}  \in \mathbb{R}^{I_{out}\times R}$. $\mathbf{\Theta}^{(l)}$ and  $\mathbf{Q}^{(l)}$ are the learnable weights in the $l$-th layer. $I_{in}$ is the input feature dimension number, ${I_{out}}$ is the output dimensionality and $R$ is the number of rank. We define the family of such hypergraph neural networks as \textbf{T}ensorized \textbf{H}ypergraph \textbf{N}eural \textbf{N}etworks~(\textbf{THNN}).

\subsection{Architecture Analysis and Model Details}
\label{aac}
Traditional GCN speeds up their computation via sparse matrix operation in Pytorch\footnote{https://pytorch.org/} or Tensorflow\footnote{https://www.tensorflow.org/}. However, sparse tensor operations have not been supported well in common differential programming libraries. The above extension would suffer from high computational space cost of huge adjacency tensor, especially when the order is high. After fully optimizing the order of calculations, we can rewrite the THNN in
\begin{align}
    &    
    {x_{v_i}}^{(l+1)}=\sum_{(j_1,j_2,\cdots,j_{N-1}) \in N_i}  (\mathbf{Q}^{(l)}(\frac{1}{(N-1)!}\frac{1}{\Pi_{l} \sqrt[N]{d_{j_l}}}\notag\\&
    \left(\mathbf{\Theta}^{(l)\top}\left[\begin{array}{c}{x}^l_{v_{j_1}} \\ 1\end{array}\right]\right)
    \star \cdots \left(\mathbf{\Theta}^{(l)\top}\left[\begin{array}{c}{x}^l_{v_{j_{N-1}}} \\ 1\end{array}\right]\right))).
    \label{Eq:ATHGCN2}
\end{align}

where $N_i$ is the set of neighbor pairs of the node $v_i$, and $\star$ is element-wise dot product.  
{
And considering that low order information can also be very important in some cases, we concatenate a scalar $1$ in feature vectors to generate lower-order dynamics. 
Such a strategy could help THNN with low-order information modeling.
In detail,} Eq.~(\ref{Eq:ATHGCN2}) considers more on
the 2nd-order interactions and ignores some 1st-order information. As for the 4-uniform situation, the 3rd-order interactions would be considered more. As shown in Figure~{\ref{fig:fusion}}, such preference can be alleviated if we concatenate original feature vector with a scalar $1$. 

As dot product of many vectors would lead the numerical insatiability empirically, we add a new activation function $\sigma^{'}$ in the original architectures. We evaluated common activation functions and used $Tanh(\cdot)$ in experiments. The final expression of THNN for uniform Hyper-Graph is represented
as follows
 \begin{align}
 \notag
 & {x_{v_i}}^{(l+1)}=\sigma(\sum_{(j_1,j_2,\cdots,j_{N-1}) \in N_i}  (\mathbf{Q}^{(l)}\textit{Tanh}(\frac{1}{(N-1)!}\\&\frac{1}{\Pi_{l} \sqrt[N]{d_{j_l}}}
    \left(\mathbf{\Theta}^{(l)\top}\left[\begin{array}{c}{x}^l_{v_{j_1}} \\ 1\end{array}\right]\right)
    \star \cdots \left(\mathbf{\Theta}^{(l)\top}\left[\begin{array}{c}{x}^l_{v_{j_{N-1}}} \\ 1\end{array}\right]\right)))).
    \label{Eq:ATHGCN3}
\end{align}
The whole procedure of THNN is shown in Figure~{\ref{fig:model}}.

\subsection{Non-Uniform Generalization}
One of the most critical issues of using adjacency tensor in hypergraph analysis is that only uniform hypergraph can be processed. In addition, the proposed THNN in Eq.~(\ref{Eq:ATHGCN3}) only considered the uniform hypergraph. But in many real-world situations, non-uniform hypergraphs are needed. Hence, we aim to extend the proposed model for non-uniform hypergraphs. Motivated by \cite{zhen2022community} and \cite{ouvrard2018adjacency}, we propose two methods to extend the uniform hypergraph models for general hypergraphs.

\noindent\textbf{Global Node.} As shown in Figure~\ref{fig:uni1}, we can add a global node to the hypergraph. The non-informative global node would be added many times in one hyperedge until the order of the hyperedge is equal to the max-order number. The feature vector of the global point will be a trainable vector with the same size as other node features. {Since the non-informative global node has too many neighbors, the representation of the linked neighbors of the informative global node will tend to converge to the same value, which may exacerbate the oversmoothing in message passing procedure.}

\noindent\textbf{Multi-Uniform Processing.} {So in order to mitigate the problem of oversmoothing of global node adding strategies, }as shown in Figure~\ref{fig:uni2}, we also proposed a multi-uniform processing scheme that decomposes a non-uniform hypergraph into several uniform sub-hypergraphs. One sub-hypergraph contains all the hyperedges with the same number of orders. Then, we could process uniform sub-hypergraphs separately. Finally, we can concatenate feature vectors with different orders and process them via a trainable weight matrix.  

{In this paper, we mainly focus on node classification tasks. Therefore, when we get the node representation through several layers of THNN, we directly use a fully connected layer to obtain the predicted label and use the \textbf{cross-entropy} loss to optimize parameters.}

\begin{figure}[t]
\centering
 \includegraphics[width=0.5\textwidth]{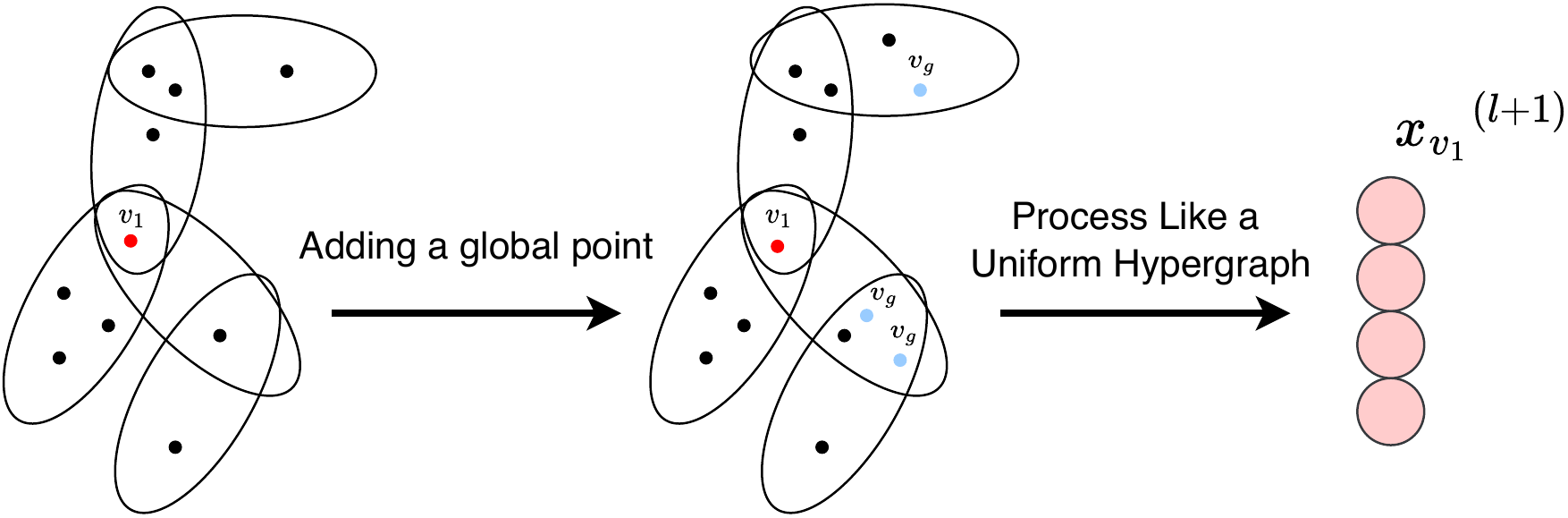}
  \caption{Inspired by~\cite{zhen2022community}, we can add a global node $v_g$ to a non-uniform hypergraph in order to make it uniform. The global node can be added many times in one hyperedge. Following this procedure, the hypergraph can well be represented as an adjacency tensor, allowing it to be directly processed in uniform-hypergraph form models. 
}
  \label{fig:uni1}
 \end{figure}

\begin{figure}[t]
\centering
 \includegraphics[width=0.5\textwidth]{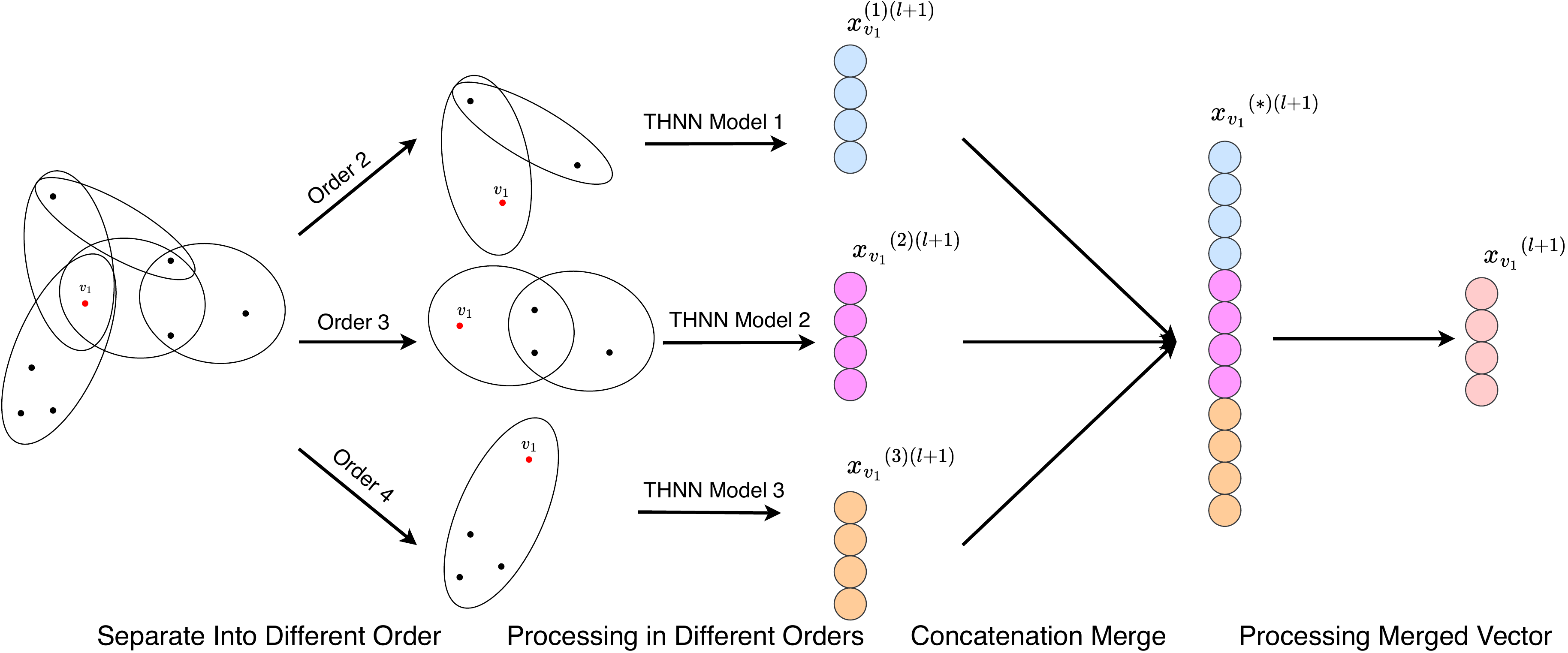}
  \caption{We can also process hypergraphs in layers and utilize distinct models to process sub-hypergraphs of different orders. The resultant embedding vectors of distinct layers are therefore concatenated and integrated through a fully-connected layer. 
  }
  \label{fig:uni2}
 \end{figure}

\section{Experiments}

In this section, we conduct experiments on two different 3-D visual object classification datasets under both uniform and non-uniform hypergraph construction settings. Experimental results verify the effectiveness of the proposed model.
{In addition, we also performed ablation analysis and hyperparameter experiments in order to acquire a better knowledge of the model. 

\subsection{Datasets }

In experiments, two public benchmarks, the Princeton ModelNet40 dataset~\cite{wu20153d} and the National Taiwan University (NTU) 3D model dataset~\cite{chen2003visual} are used. The ModelNet40 dataset includes 12,311 objects from 40 popular categories, while the NTU dataset has 2,012 3D items from 67 categories. We apply the similar data split settings in~\cite{feng2019hypergraph, ran2022surface}, and we extract the features of 3D objects through the Multi-view Convolutional Neural Network~(MVCNN)~\cite{su2015multi} and Group-view Convolutional Neural Network~(GVCNN)~\cite{feng2018gvcnn}. 12 virtual cameras are used to collect images with a 30-degree interval angle, and MVCNN features and GVCNN features are then extracted accordingly.

\subsection{Hypergraph Generation}

After obtaining the vector embedding representation of the 3D object in Euclidean space, we implement a distance-based hypergraph generation~\cite{gao20123,feng2019hypergraph,gao2020hypergraph}. Such a distance-generation
approach connects a group of $k$ similar vertices to the same centroid and exploits the correlations among vertices. Distance-based hyperedges can represent node connection in feature space~\cite{gao2020hypergraph}.

\subsection{Baseline Models}

After constructing hypergraphs, we compare THNN with multiple graph and hypergraph neural network baselines. All results about baseline are reproduced by ourselves. For GNN baselines, we feed the clique-expansion of constructed hypergraphs. The baselines are listed and decribed as follows: Graph convolutional network~(GCN)~\cite{welling2016semi}, Graph attention network~(GAT)~\cite{velickovic2017graph}, Graph Isomorphism Network~(GIN)~\cite{leskovec2019powerful},HyperGCN~\cite{yadati2019hypergcn}, Hypergraph Networks with Hyperedge Neurons~(HNHN)~\cite{dong2020hnhn}, Hypergraph Neural Networks~(HGNN)~\cite{feng2019hypergraph}.

\subsection{Results on the Uniform Setting}
First, we evaluate THNN along with the baselines in a uniform hypergraph setting with $K=4$. We employ a two-layer THNN model with a rank setting of 128. THNN is trained with the Adam optimizer whose learning rate is set to be 0.001 initially.
Similar to the evaluation procedure in~\cite{feng2019hypergraph}, we create multiple hypergraph structures for comparison using either the single features or the concatenation multi-features. 
Detailed results on the two datasets are reported in Tables~\ref{tab:uni}. 

We have the following observations:
First, THNN maintains the best performance in the majority of cases because the proposed models are compelling in extracting high-order information of hypergraph structures. For example, compared with HyperGCN, THNN achieves gains of $0.74\%$ and $1.03\%$ on average of 7 settings on the ModelNet40 and the NTU datasets, respectively.
Second, the graph-based model performs worse than the hypergraph in most cases because the hypergraph structure can convey complicated high-order correlations among data, whereas the clique expansion introduces some information loss.
Furthermore, as a result of information loss, the graph-based models achieve poor results in some settings. For example, GCN achieves $77.72\%$, and GIN achieves $88.72\%$ in the \textbf{C: MvGv, T: Mv} setting of ModelNet40, while the mean accuracy of other hypergraph models under such setting is all above $90\%$. The representation information under this setting can be better characterized by a model considering higher-order information.
These cases indicate that simple graph-based models are unstable when higher-order information dominates.

\begin{table*}[]
\centering
\caption{Experiment result of uniform generation setting. ``C" means the feature type used in hypergraph construction. "T" the feature type used in model training (as the node feature vector). ``Mv" means the MVCNN feature, ``Gv" means the GVCNN feature and ``MvGv" means the concatenation of MVCNN feature and GVCNN feature. We run 5 times in each setting and the results are displayed in (mean $\pm$ std) form.} 
\label{tab:uni}
\resizebox{\textwidth}{!}{%
\begin{tabular}{c|lllllll|lllllll}
\toprule
\toprule
{Dataset} & \multicolumn{7}{c|}{{NUT2012}} & \multicolumn{7}{c}{{ModelNet40}} \\ \hline
{Setting} & {\begin{tabular}[c]{@{}l@{}}C: Mv \\ T: Mv\end{tabular}} & {\begin{tabular}[c]{@{}l@{}}C: Mv \\ T: Gv\end{tabular}} & {\begin{tabular}[c]{@{}l@{}}C: Gv \\ T: Mv\end{tabular}} & {\begin{tabular}[c]{@{}l@{}}C: Gv \\ T: Gv\end{tabular}} & {\begin{tabular}[c]{@{}l@{}}C: MvGv \\ T: Mv\end{tabular}} & {\begin{tabular}[c]{@{}l@{}}C: MvGv \\ T: Gv\end{tabular}} & {\begin{tabular}[c]{@{}l@{}}C: MvGv \\ T: MvGv\end{tabular}}  & {\begin{tabular}[c]{@{}l@{}}C: Mv \\ T: Mv\end{tabular}} & {\begin{tabular}[c]{@{}l@{}}C: Mv \\ T: Gv\end{tabular}} & {\begin{tabular}[c]{@{}l@{}}C: Gv \\ T: Mv\end{tabular}} & {\begin{tabular}[c]{@{}l@{}}C: Gv \\ T: Gv\end{tabular}} & {\begin{tabular}[c]{@{}l@{}}C: MvGv \\ T: Mv\end{tabular}} & {\begin{tabular}[c]{@{}l@{}}C: MvGv \\ T: Gv\end{tabular}} & {\begin{tabular}[c]{@{}l@{}}C: MvGv \\ T: MvGv\end{tabular}}\\ \hline
GCN & 71.31$\pm$2.54\% & 75.23$\pm$2.11\%& 74.23$\pm$2.07\%& 78.76$\pm$2.14\%& 80.37$\pm$2.28\%& 80.84$\pm$1.93\% & 79.57$\pm$1.05\% & 85.12$\pm$1.75\% & 89.57$\pm$1.38\% & 83.09$\pm$1.23\% & 89.91$\pm$1.61\% & 86.80$\pm$3.05\% & 91.97$\pm$2.89\% & 88.21$\pm$3.03\% \\
GAT & 74.45$\pm$0.69\%& 77.81$\pm$0.92\% & 81.31$\pm$0.91\% & 82.94$\pm$0.58\% & 82.57$\pm$0.28\% & 83.11$\pm$0.13\% & 80.11$\pm$0.67\% & 86.41$\pm$0.84\% & 90.79$\pm$0.81\% & 88.99$\pm$0.98\% & 91.55$\pm$0.21\% & 89.86$\pm$0.53\% & 95.03$\pm$0.55\% & 93.50$\pm$0.18\% \\ 
GIN & 75.71$\pm$1.13\% & 78.38$\pm$0.78\% & 78.39$\pm$0.73\% & 80.77$\pm$0.85\% & 80.33$\pm$0.98\% & 82.47$\pm$0.28\% & 78.75$\pm$2.03\% & 86.74$\pm$0.57\% & 91.57$\pm$0.43\% & 89.97$\pm$0.43\% & 91.39$\pm$0.18\% & 88.79$\pm$1.38\% & 91.86$\pm$0.76\% & 91.95$\pm$0.32\%  \\ 
HyperGCN & 77.21$\pm$0.35\% & {78.81$\pm$0.22\%} & 83.95$\pm$0.28\% & {84.03$\pm$0.17\%} & 80.09$\pm$0.12\% & 81.37$\pm$0.45\% & 80.32$\pm$0.76\% & {88.52$\pm$0.43}\% & 91.04$\pm$0.98\% & 91.01$\pm$0.28\% & 91.87$\pm$0.24\% & 90.02$\pm$0.34\% & 95.08$\pm$0.16\% & 92.01$\pm$0.17\% \\ 
HNHH & 77.33$\pm$0.15\% & {78.88$\pm$0.92\%} & 84.03$\pm$0.41\% & 81.91$\pm$0.32\% & 81.94$\pm$0.73\% & {82.72$\pm$0.37}\% & 82.24$\pm$0.33\% & 86.56$\pm$0.34\%& 91.15$\pm$0.18\% & 90.05$\pm$0.97\% & 91.93$\pm$0.58\% & 91.35$\pm$0.27\% & 94.36$\pm$0.22\% & 93.01$\pm$0.36\% \\ 
HGNN & {77.27$\pm$0.12\%} & 78.02$\pm$1.06\% & 81.65$\pm$0.23\% & {83.98$\pm$0.21}\% & 81.79$\pm$0.38\% & 82.67$\pm$0.19\% & 82.87$\pm$0.28\% & 86.79$\pm$0.34\% & 91.96$\pm$0.42\% & 91.91$\pm$0.18\% & 91.76$\pm$0.23\% & 91.72$\pm$0.11\% & 95.75$\pm$0.13\% & 92.37$\pm$0.27\% \\ \hline
THNN & \textbf{77.33$\pm$0.20\%} & 78.21$\pm$0.27\% & {84.21$\pm$0.15\%} & {83.98$\pm$0.15}\% & {82.71$\pm$0.32}\% & {82.77$\pm$0.24}\% & {83.71$\pm$0.22}\% & 87.55$\pm$0.25\% & {92.18$\pm$0.32\%} & {91.91$\pm$0.21\%} & {92.02$\pm$0.18}\% & {92.53$\pm$0.19}\% & {96.11$\pm$0.07\%
} & {93.65$\pm$0.15}\% \\ 
\bottomrule
\bottomrule
\end{tabular}%
}
\end{table*}

\begin{table*}[]
\centering
\caption{{Experiment result of \textbf{non}-uniform hypergraph. Compared with uniform hypergraph results in Table~\ref{tab:uni}, the hypergraph structure is non-uniform in this table. The original THNN is incapable of processing non-uniform hypergraph structures. So, we report the result of two non-uniform extensions of THNN in this table. The {non}-uniform hypergraph is generated via Bernoulli sampling on the probabilistic incidence matrix and the hypergraph structure is the same in one setting. }}
\label{tab:non-uni}
\resizebox{\textwidth}{!}{%
\begin{tabular}{c|lllllll|lllllll}
\toprule
\toprule
\textbf{Dataset} & \multicolumn{7}{c|}{\textbf{NUT2012}} & \multicolumn{7}{c}{\textbf{ModelNet40}} \\ \hline
\textbf{Setting} & \textbf{\begin{tabular}[c]{@{}l@{}}C: Mv \\ T: Mv\end{tabular}} & \textbf{\begin{tabular}[c]{@{}l@{}}C: Mv \\ T: Gv\end{tabular}} & \textbf{\begin{tabular}[c]{@{}l@{}}C: Gv \\ T: Mv\end{tabular}} & \textbf{\begin{tabular}[c]{@{}l@{}}C: Gv \\ T: Gv\end{tabular}} & \textbf{\begin{tabular}[c]{@{}l@{}}C: MvGv \\ T: Mv\end{tabular}} & \textbf{\begin{tabular}[c]{@{}l@{}}C: MvGv \\ T: Gv\end{tabular}} & \textbf{\begin{tabular}[c]{@{}l@{}}C: MvGv \\ T: MvGv\end{tabular}}  & \textbf{\begin{tabular}[c]{@{}l@{}}C: Mv \\ T: Mv\end{tabular}} & \textbf{\begin{tabular}[c]{@{}l@{}}C: Mv \\ T: Gv\end{tabular}} & \textbf{\begin{tabular}[c]{@{}l@{}}C: Gv \\ T: Mv\end{tabular}} & \textbf{\begin{tabular}[c]{@{}l@{}}C: Gv \\ T: Gv\end{tabular}} & \textbf{\begin{tabular}[c]{@{}l@{}}C: MvGv \\ T: Mv\end{tabular}} & \textbf{\begin{tabular}[c]{@{}l@{}}C: MvGv \\ T: Gv\end{tabular}} & \textbf{\begin{tabular}[c]{@{}l@{}}C: MvGv \\ T: MvGv\end{tabular}}
\\ \hline
GCN & 71.09$\pm$1.04\% & 74.01$\pm$1.01\% & 73.51$\pm$3.59\% & 77.73$\pm$3.63\% & 80.23$\pm$0.57\% & 80.01$\pm$0.91\% & 81.17$\pm$0.83\% & 84.73$\pm$0.98\% & 91.77$\pm$0.67\% & 77.95$\pm$3.21\% & 90.10$\pm$1.35\% & 77.72$\pm$1.53\% & 94.47$\pm$1.09\% & 88.79$\pm$2.29\% \\ 
GAT & 75.04$\pm$0.87\% & 76.45$\pm$0.76\% & 80.56$\pm$0.82\% & 80.83$\pm$0.37\% & 79.67$\pm$0.62\% & 81.34$\pm$0.54\% & 77.62$\pm$0.78\% & 84.62$\pm$2.14\% & 89.65$\pm$1.29\% & 88.04$\pm$0.92\% & 90.17$\pm$0.63\% & 88.24$\pm$0.49\% & 94.52$\pm$0.57\% & 92.45$\pm$0.72\%   \\ 
GIN & 75.53$\pm$0.43\% & 75.37$\pm$0.99\% & 79.17$\pm$0.84\% & 80.07$\pm$0.66\% & 77.88$\pm$1.83\% & 82.32$\pm$0.24\% & 77.57$\pm$1.08\% & 87.23$\pm$0.59\% & 91.42$\pm$0.97\% & 87.03$\pm$1.32\% & 91.17$\pm$0.43\% & 88.72$\pm$1.50\% & 92.01$\pm$1.16\% & 94.29$\pm$0.65\% \\ 
HyperGCN & 75.88$\pm$0.91\% & 76.72$\pm$0.32\% & 80.67$\pm$0.65\% & 82.23$\pm$1.07\% & 79.68$\pm$0.79\% & 81.63$\pm$0.91\% & 80.57$\pm$1.06\% & 87.59$\pm$0.95\% & 93.53$\pm$0.14\% & 91.91$\pm$0.13\% & 91.26$\pm$0.89\% & 91.19$\pm$1.08\% & 95.43$\pm$0.53\% & 93.72$\pm$0.61\% \\ 
HNHH & 75.06$\pm$0.69\% & 76.41$\pm$0.36\% & 81.29$\pm$0.69\% & 81.70$\pm$1.32\% & 82.79$\pm$0.63\% & 82.76$\pm$0.65\% & 81.54$\pm$0.76\% & 88.28$\pm$0.34\% & 90.25$\pm$1.54\% & 91.76$\pm$0.33\% & 91.05$\pm$0.97\% & 90.79$\pm$0.82\% & 94.99$\pm$0.32\% & 93.75$\pm$0.99\% \\ 
HGNN & 76.06$\pm$0.87\% & 77.25$\pm$0.15\% & 83.01$\pm$0.53\% & 81.57$\pm$0.47\% & 82.04$\pm$0.61\% & 83.01$\pm$0.96\% & 82.09$\pm$0.83\% & 89.33$\pm$0.78\% & 93.01$\pm$0.92\% & 91.13$\pm$0.54\% & 91.05$\pm$0.37\% & 90.32$\pm$1.03\% & 95.71$\pm$0.16\% & 94.91$\pm$0.12\%  \\ \hline
THNN-AdG & 74.34$\pm$0.92\% & 77.01$\pm$0.19\% & 83.38$\pm$0.41\% & 82.01$\pm$0.11\% & 83.08$\pm$0.76\% & 81.50$\pm$0.87\% & 82.04$\pm$0.35\% & 87.44$\pm$1.01\% & 92.16$\pm$0.83\% & 91.31$\pm$0.56\% & 92.12$\pm$0.32\%& 93.01$\pm$0.81\% & 96.05$\pm$0.17\% & 93.33$\pm$0.32\% \\ 
\multicolumn{1}{l|}{THNN-Multi} & 77.05$\pm$0.26\% & 77.42$\pm$0.14\% & 82.07$\pm$0.82\% & 82.84$\pm$0.06\% & 83.79$\pm$0.21\% & 82.09$\pm$0.27\% & 82.27$\pm$0.34\% & 90.01$\pm$0.31\% & 94.23$\pm$0.28\% & 91.42$\pm$0.37\% & 91.61$\pm$0.93\% & 93.19$\pm$0.73\% & 95.93$\pm$0.12\% & 95.07$\pm$0.13\%  \\ 
\bottomrule
\bottomrule
\end{tabular}%
}
\end{table*}

\subsection{Results on the non-Uniform Setting}
We also create non-uniform hypergraph structures to validate the efficacy of the two proposed extensions of THNN. Because each element of the probability incidence matrix ${\Hat{H}}$ takes value in $[0, 1]$, and the closer the value is to 1, the more similar with centroid node the value is, it is possible to perform Bernoulli sampling on ${\Hat{H}}$. We employ the two-layer THNN extension models with adding a global node~(THNN-AdG) and Multi-Uniform Processing~(THNN-Multi) a rank setting of 128. The two extensions are also trained with Adam optimizers whose learning rates are set to be 0.005.

Detailed results are reported in Tables~\ref{tab:non-uni}.
In these two tables, the results remain consistent with those in the uniform settings.
The two proposed extensions of THNN perform the best in the majority of scenarios.
Specifically, THNN-Multi performs better than THNN-AdG.
Compared with THNN-AdG, THNN-Multi achieves gains of $1.18\%$ and $0.76\%$ on average of the 7 settings on the ModelNet40 and the NTU datasets, respectively.
The possible reason is that interactions of different orders are learned via hypergraph separation processing procedure and aligned via merged layer in THNN-Multi, while artificially introduced external global node might disturb the interaction representation of different orders {due to the oversmoothing issue} in THNN-AdG. 
But THNN-Multi requires several THNN models to process the hypergraph message passing procedure of different orders. 
As a result, the number of parameters of THNN-Multi increases linearly with the order number, but the parameter number of the THNN-AdG is not influenced by order number.
Thus, the well-performing THNN-Multi generally has a larger number of parameters than THNN-AdG when the order number is large.
In addition, the graph-based model still performs worse than the hypergraph in most cases and suffers from large variance across hypergraph construction and training feature selection settings.
{In summary, hypergraph-based models can better model complex correlations among data than graph-based models. Compared with other hypergraph models that adopt first-order approximations,
our proposed adjacency-tensor-based THNN enjoys higher-order information modeling, leading to better performance.
}

\subsection{Ablation Study}

We also conducted some ablation experiments to verify the effectiveness of the proposed techniques. {We choose two uniform hypergraph generation settings.} For these two datasets, we evaluate the proposed three techniques. We use $Tanh$ in $\sigma'$ to make the element-wise product in CP decomposition stable. Secondly, we use the techniques of the \textit{concatenating ones} 
to improve the expressiveness of the model. We also evaluate the normalized adjacency tensor strategy, which is inspired by the normalized Laplacian adjacency matrix to make the training stable. Results show that it is crucial to use $Tanh$. This is because higher-order tensor operations introduce operations such as $x ^ n $, which makes training neural networks extremely challenging. Small changes in the $x$ can cause large disturbances in $x ^ n $ function.
\begin{table}[]
\setlength{\belowcaptionskip}{-0.5cm} 
\centering
\caption{Ablation Experiment Results. Results show that it is crucial to use the activation function like $Tanh$, and other techniques can also improve the performance. }
\label{tab:my-table}
\resizebox{.95\columnwidth}{!}{%
\begin{tabular}{c|ll|ll}
\toprule
\toprule
\textbf{Dataset} & \multicolumn{2}{c|}{NUT2012} & \multicolumn{2}{c}{ModelNet40} \\ \hline
\textbf{Setting} & \textbf{\begin{tabular}[c]{@{}l@{}}C: Mv\\ T: Gv\end{tabular}} & \textbf{\begin{tabular}[c]{@{}l@{}}C: MvGv\\ T: MvGv\end{tabular}} & \textbf{\begin{tabular}[c]{@{}l@{}}C: Mv\\ T: Gv\end{tabular}} & \textbf{\begin{tabular}[c]{@{}l@{}}C: MvGv\\ T: MvGv\end{tabular}} \\ \hline
No Tanh & 69.43\% & 68.36\% & 82.74\% & 83.95\% \\ 
No Concatenating Ones & 77.47\% & 82.84\% & 92.14\% & 92.26\% \\ 
No Degree Normalization & 77.47\% & 82.57\% & 92.38\% & 92.38\% \\ \hline
THNN & 78.55\% & 83.91\% & 92.38\% & 94.25\% \\ 
\bottomrule
\bottomrule
\end{tabular}%
}
\end{table}

\subsection{Hyper-parameter Analysis: Rank and Number of Layers}

\label{Sec:Hyper}
As shown in Fig~\ref{fig:rank}, we evaluate the impact of rank $R$ on model performance by generating hypergraphs with varying values of $K$ (from 2 to 6) in the NUT2012 experiment and by adjusting the rank setting in the THNN. We choose the "C: Mv T: Gv" setting in NUT2012. We found that $Rank = 128$ is a proper setting, as the accuracy does not change much when $Rank \in [75,150]$. We also examine how the number of layers in THNN affects performance. Under the setting of "C: Mv T: Gv" in NUT2012, we discover that the model's expressiveness cannot be fully explored with a single layer. Still, the model's performance degrades dramatically when there are too many layers. We conject that this may be due to numerical instability (for example, $10.001^{10}$ is much larger than $10^{10}$) of higher order operations or over-smoothing problems in graph learning~\cite{chen2020measuring}. The experiment results in Fig~\ref{fig:layer} has demonstrated that stacking two layers is optimal.

\begin{figure}[!htbp]
     \centering
     \begin{subfigure}[b]{0.23\textwidth}
         \centering
         \includegraphics[width=\textwidth]{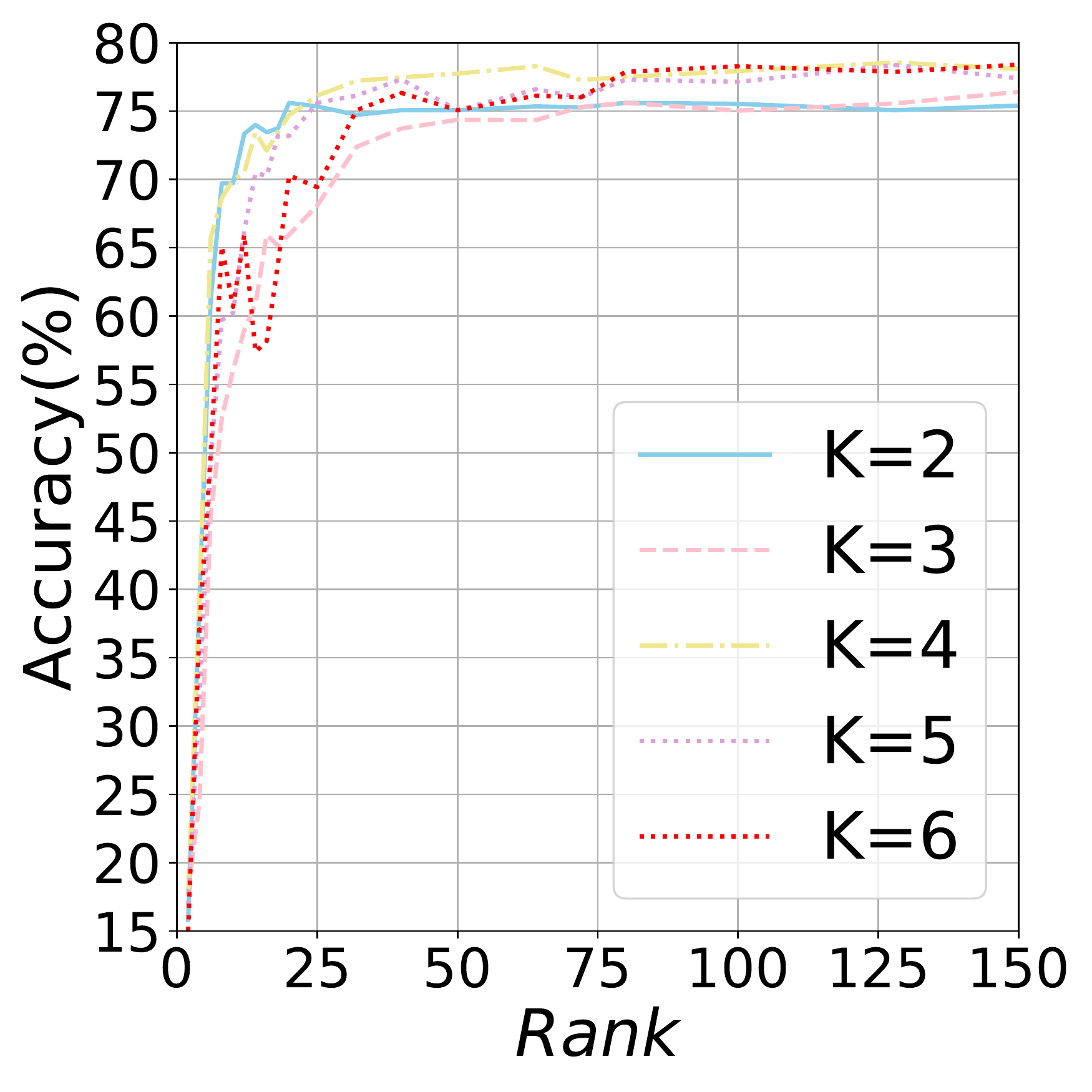}
         \caption{Rank analysis result.}
         \label{fig:rank}
     \end{subfigure}
     \hfill
          \begin{subfigure}[b]{0.24\textwidth}
         \centering
         \includegraphics[width=\textwidth]{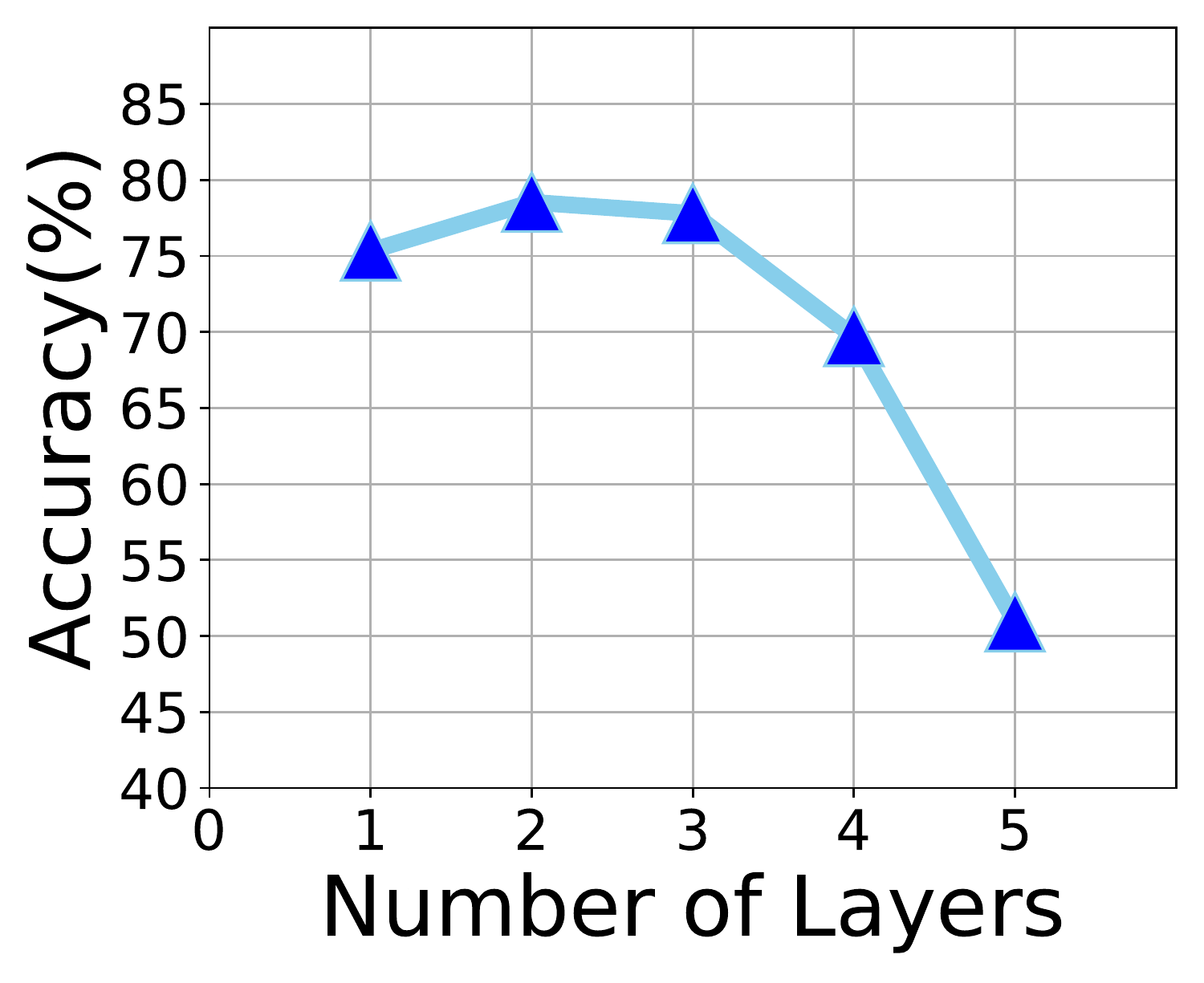}
         \caption{Influence of number layers.}
         \label{fig:layer}
     \end{subfigure}
     \caption{Hyper-parameter analysis. (a) $Rank = 128$ is a proper setting, as the accuracy does not change much when $Rank \in [75,150]$. 
 (b) The performance of THNN degrades dramatically when there are too many layers.}
 \end{figure}

\section{Related Works}
\subsection{Hypergraph Learning with Adjacency Tensor} 
In hypergraph learning, most existing methods focus on converting the hypergraph into a weighted graph~\cite{yadati2019hypergcn,dong2020hnhn,feng2019hypergraph}, thereby allowing the application of traditional graph methods. For instance, spectral clustering can be performed on the weighted graph Laplacian~\cite{ghoshdastidar2017consistency} to detect communities within hypergraphs. However, as highlighted in~\cite{ke2019community}, this conversion process can lead to significant information loss and consequently, sub-optimal performance in detecting hypergraph communities. Given that tensors naturally represent high-order relationships~\cite{ouvrard2018adjacency}, several methods based on adjacency tensors have been proposed. These methods aim to achieve optimal community detection~\cite{ke2019community} in both uniform and non-uniform hypergraphs. Despite the majority of adjacency tensor-based research focusing on the spectral properties~\cite{xie2013z,pearson2014spectral,pearson2013eigenvalues} of hypergraphs and tensor decomposition modelings~\cite{nickel2013tensor,zhen2022community}, there appears to be a lack of hypergraph neural networks based on adjacency tensors. This gap in the literature presents an opportunity for significant advancements in this field.

\subsection{Tensor Fusion Models and Tensorized Neural Networks} 

Zadeh et al. \cite{zadeh2017tensor} introduced a pioneering concept known as the Tensor Fusion Layer (TFL), which utilizes a tensorized outer product to serve as a deep information fusion layer. The TFL framework is designed to learn both intra-modality and inter-modality dynamics while efficiently aggregating interactions from multiple modalities. Nevertheless, the TFL model faces challenges related to computational complexity and an increase in the number of parameters, particularly when the number of modalities grows.
In response to the scalability concerns of TFL, the Low-rank Multimodal Fusion (LMF)\cite{liu2018efficient} method was proposed. LMF employs low-rank tensor decomposition techniques to reduce the computational burden associated with TFL. This type of fusion operation can be regarded as 
a specialized form of Higher-order Polynomial Regression~\cite{hsu2010microstructural, cheng1998polynomial}.
The proposed THNN represents the first Tensorized Neural Network tailored to model high-order information interactions within hypergraphs.

\section{Conclusion and Discussion}
In this paper, we discover that existing hypergraph neural networks are mostly based on the message passing of first-order approximations of the original hypergraph structure and ignore higher-order interactions encoded in hypergraphs. In order to better model the higher-order information of the hypergraph structure, we propose a novel hypergraph neural network, THNN, which is based on adjacency tensors of hypergraphs and is a high-order extension of traditional Graph Convolution Neural Networks.
We also find that the proposed models are
well connected with tensor fusion, which is a highly recognized successful technique for high-order interaction modeling. Therefore, the proposed models are compelling in extracting high-order information encoded in hypergraphs.
We show that our framework can achieve the promising performance in 3-D visual object classification tasks under both uniform and non-uniform hypergraph settings.
In the future, we plan to explore more applications of THNN to verify its advantages.

\section{Acknowledge}
This research was partially supported by Research Impact Fund (No.R1015-23), APRC - CityU New Research Initiatives (No.9610565, Start-up Grant for New Faculty of City University of Hong Kong), CityU - HKIDS Early Career Research Grant (No.9360163), Hong Kong ITC Innovation and Technology Fund Midstream Research Programme for Universities Project (No.ITS/034/22MS), Hong Kong Environmental and Conservation Fund (No.88/2022), and SIRG - CityU Strategic Interdisciplinary Research Grant (No.7020046, No.7020074), Ant Group (CCF-Ant Research Fund, Ant Group Research Fund), Huawei (Huawei Innovation Research Program), Tencent (CCF-Tencent Open Fund, Tencent Rhino-Bird Focused Research Program), CCF-BaiChuan-Ebtech Foundation Model Fund, and Kuaishou.

\newpage
\bibliographystyle{siam}
\bibliography{mybib}
\end{document}